\documentclass{article}

\usepackage{tikz}

\usepackage{microtype}
\usepackage{graphicx}
\usepackage{subfigure}
\usepackage{booktabs} 
\usepackage{amsmath}
\usepackage{amssymb}

\usepackage{hyperref}

\renewcommand{\xi}[1][i]{\mathbf{x}^{(#1)}}                                          

\usepackage{pxfonts}
\usepackage{mathtools}

\newcommand{\idp}{\Perp}

\newtheorem{assumption}{Assumption}
\newtheorem{definition}{Definition}
\newtheorem{proposition}{Proposition}

\usepackage[accepted]{icml2021}

\icmltitlerunning{A Causal Perspective on Meaningful and Robust Algorithmic Recourse}

\begin{document}

\twocolumn[
\icmltitle{A Causal Perspective on Meaningful and Robust Algorithmic Recourse}




\begin{icmlauthorlist}
\icmlauthor{Gunnar König}{lmu,uvienna}
\icmlauthor{Timo Freiesleben}{mcmp}
\icmlauthor{Moritz Grosse-Wentrup}{uvienna}
\end{icmlauthorlist}

\icmlaffiliation{lmu}{Institute for Statistics, LMU Munich}
\icmlaffiliation{uvienna}{Research Group Neuroinformatics, University of Vienna}
\icmlaffiliation{mcmp}{Munich Center for Mathematical Philosophy, LMU Munich}

\icmlcorrespondingauthor{Gunnar König}{g.koenig.edu@pm.me}

\icmlkeywords{algorithmic recourse, causality, counterfactual explanations, robustness}

\vskip 0.3in
]



\printAffiliationsAndNotice{}  

\begin{abstract}
Algorithmic recourse explanations inform stakeholders on how to act to revert unfavorable predictions.
However, in general ML models do not predict well in interventional distributions. Thus, an action that changes the prediction in the desired way may not lead to an improvement of the underlying target.
Such recourse is neither meaningful nor robust to model refits.
Extending the work of \citet{karimi_algorithmic_2021}, we propose meaningful algorithmic recourse (MAR) that only recommends actions that improve both prediction and target. We justify this selection constraint by highlighting the differences between model audit and meaningful, actionable recourse explanations.
Additionally, we introduce a relaxation of MAR called effective algorithmic recourse (EAR), which, under certain assumptions, yields meaningful recourse by only allowing interventions on causes of the target.
\end{abstract}

\section{Introduction}
\label{sec:introduction}

\begin{figure}[H]
    \centering
    \begin{tikzpicture}[thick, scale=0.6, every node/.style={scale=.65, line width=0.25mm, black, fill=white}]
    \usetikzlibrary{shapes}
		\node[draw=green, ellipse, scale=0.9] (x1) at (-1.5, 1) {car owner};
		\node[draw=green, dotted, ellipse, scale=0.9] (x2) at (1.5, 1) {defensiveness};
		\node[draw=blue, ellipse, scale=0.9] (x3) at (3, 0) {minivan};
		\node[draw, ellipse, scale=0.9] (y) at (0,0) {$Y:$ insurance risk};
		\draw[->] (x1) -- (y);
		\draw[->] (x2) -- (y);
		\draw[->] (x2) -- (x3);
		
		\draw[-, dotted] (4.5, 1.5) -- (4.5, -.5);
		
		\node[draw=green, ellipse, scale=0.9] (x1) at (6, 1) {car owner};
		\node[draw=green, dotted, ellipse, scale=0.9] (x2) at (9, 1) {defensiveness};
		\node[draw=blue, ellipse, scale=0.9] (x3) at (10.5, 0) {minivan};
		\node[draw, ellipse, scale=0.9] (yh) at (7.5,0) {$\hat{Y}:$ prediction};
		\draw[->] (x1) -- (yh);
		\draw[->] (x3) -- (yh);
    \end{tikzpicture}
    \caption{Left: Directed Acyclic Graph (DAG) illustrating the data generating process. Right: Directed Acyclic Graph (DAG) illustrating prediction model.}
    \label{fig:example-intro}
\end{figure}
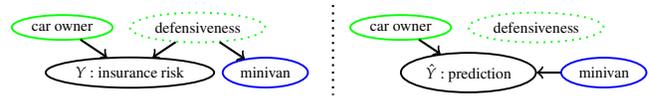

Predictive systems are increasingly deployed in high-stakes environments such as hiring \cite{Manish2020}, recidivism prediction \cite{zeng2015interpretable} or loan approval \cite{van2017machine}. To enable individuals to revert unfavorable decisions, a range of work develops tools that offer individuals possibilities for algorithmic recourse \cite{Wachter2018,dandl_multi-objective_2020,karimi_algorithmic_2020,karimi_algorithmic_2021}. When suggesting actions for recourse, it is desirable that recommendations are \textit{robust} to shifts, meaning that they can be honored if acted upon \cite{venkatasubramanian_philosophical_2020}. We argue that they should also be \textit{meaningful}, such that not only the prediction, but also the underlying target is improved.\\
We take a causal perspective on the issue at hand and argue that robustness and meaningfulness are related problems since non-meaningful recourse itself leads to distribution shift. 
Let us consider a simple motivational example\footnote{The example is inspired by \cite{shavit_causal_2020}.} illustrated in Figure \ref{fig:example-intro}. The goal is to predict the insurance risk of a car. In addition to the direct cause of whether the car is driven by the \textit{car owner} (green), the confounded variable \textit{minivan} is observed (blue). The latent confounder driver \textit{defensiveness} cannot be observed. The ML model learns to exploit not only the direct cause (green) but also the associated variable (blue). Algorithmic recourse actions on the model may therefore suggest explainees to game the predictor by intervening on the non-causal variable minivan, thereby affecting the prediction without actually improving the insurance risk. 
In the distribution of agents that have gamed the prediction model, the association of \textit{minivan} with the prediction target is broken. A refitted model would adapt accordingly and the recourse would not be honored.
\citet{miller_strategic_2020} suggest to adapt the model such that gaming is not incentivized, which may come at the cost of predictive performance \cite{shavit_causal_2020}. Instead we suggest to tackle the problem in the explanation domain. We propose Meaningful Algorithmic Recourse (MAR) (Section \ref{sec:mar}), which restricts recourse recommendations to meaningful actions that alter both the prediction $\hat{Y}$ and the target $Y$ coherently. We justify the restriction by separating two goals: contestability and recourse (Section \ref{sec:two-tales}). While stakeholders that seek explanations for model audit (contestability) should be given full access to the model, agents that seek to revert an unfavorable outcome (recourse) should only be offered recourse options that are meaningful. A relaxation of the meaningfulness restriction is introduced in Section \ref{sec:robustness-mar}.
\section{Related Work and Contributions}
\label{sec:related-work}
\textbf{Causal Perspective on Strategic Modeling:} The related field of strategic modeling investigates how the prediction mechanism incentivizes rational agents. \citet{miller_strategic_2020} thereby distinguish models that incentivize \textit{gaming}, i.e., interventions that affect the prediction $\hat{Y}$ but not the underlying target $Y$ in the desired way, and \textit{improvement}, actions that yield the desired change both in $\hat{Y}$ and $Y$.
Building on this distinction, \cite{shavit_causal_2020} elaborate that except for cases where all causes can be measured, the following three goals are in conflict: incentivizing improvement, predictive accuracy and retrieving the true underlying mechanism.\\ 
\textbf{Robust Algorithmic Recourse:} \citet{barocas_hidden_2020,venkatasubramanian_philosophical_2020} argue that counterfactual explanations (CE) assume the model to be stable over time, but that recourse should be guaranteed even if the model changes. In a similar vein, \citet{Wachter2018} suggest guaranteeing recourse based on a counterfactual within a pre-specified period of time.
The robustness of CE and recourse has been investigated before \cite{rawal_can_2020,upadhyay_towards_2021,pawelczyk_counterfactual_2020}, yet only with respect to generic shifts. To the best of our knowledge, we are the first to suggest recommendations that are robust to the shift induced by recourse itself.\\
\textbf{Contributions:} We suggest to restrict algorithmic recourse to recommend meaningful actions that improve prediction and outcome coherently. In contrast to work in the strategic modeling literature \cite{miller_strategic_2020}, our approach does not require to adapt the model at the cost of predictive performance. We justify the restriction by distinguishing two explanation goals: model audit and algorithmic recourse. Furthermore, we suggest a relaxation of MAR that does not require computing the structural counterfactual for $Y$. We derive assumptions given which the relaxation guarantees meaningful recourse.
\section{Background and Notation}
\label{sec:background-notation}
\subsection{Causal Data Model}
We model the data generating process using a structural causal model (SCM) $\mathcal{M} \in \Pi$ \cite{Pearl2009,Peters2017book}. The model $\mathcal{M} = \langle \mathbb{F}, X, U \rangle$ consists of the endogenous variables $X \in \mathcal{X}$, the mutually independent exogenous variables $U \in \mathcal{U}$ and a sequence of structural equations $\mathbb{F}: \mathcal{U} \to \mathcal{X}$. The SCM entails a directed graph $\mathcal{G}$. The structural equations specify how $X$ is determined from $U$. The index set of endogenous variables is denoted as $D$, the set of observed variables as $O \subseteq D$. 
The Markov blanket $MB_O(Y)$ is the minimal subset of $O$, such that $X_O \idp Y | MB_O(Y)$.\footnote{Sometimes the MB is defined as minimal $d$-separating set.}
\subsection{Actionable Recourse}
Following \cite{karimi_algorithmic_2021}, we model actions as structural interventions $a : \Pi \to \Pi$. In this framework, an action can be constructed as $a = do(\{X_i := a_i\}_{i \in I})$ where $I$ is the index set of features to be intervened upon.\\
Assuming invertability of $\mathbb{F}$, the effect of an intervention for an individual can be determined using structural counterfactuals that are computed in three steps \cite{Pearl2009}: First, the factual distribution of exogenous variables given the endogenous variables is computed (abduction), i.e., $U^F = \mathbb{F}^{-1}(x^F)$. Second, the structural interventions corresponding to $a$, yielding $\mathbb{F}_a$ are performed (action). Finally, the counterfactuals are predicted $x^{SCF} = \mathbb{F}_a(u^F)$.\\
The optimization problem for recourse through minimal interventions \cite{karimi_algorithmic_2021} is given by
$$a^* \in \underset{a \in \mathbb{A}}{\text{argmin}} \text{    cost}(a, x^F) \text{  subject to } f(x^{SCF,a}) \geq t$$ 
where $\mathbb{A}$ is the action space, $x^{SCF,a} = \mathbb{F}_{a}(\mathbb{F}^{-1}(x^F))$, $f$ is the predictor and $t$ a threshold. Further constraints have been suggested, e.g., $x^{SCF,a} \in \mathcal{P}\text{lausible}$ or $a \in \mathcal{F}\text{easible}$.
\subsection{Generalizability and Intervention Stability}
\label{sec:background-stability}
We leverage necessary conditions for invariant conditional distributions as derived in \cite{pfister_stabilizing_2019}. The authors introduce a $d$-separation based intervention stability criterion that is applied to modified version of $\mathcal{G}$. For every intervened upon variable $X_l$ an auxiliary intervention variable, denoted as $\mathcal{I}_l$, is added as direct cause of $X_l$, yielding $\mathcal{G}^*$. The intervention variable can be seen as a switch between different mechanisms. A set $S \subseteq \{1, \dots, d\}$ is called \textit{intervention stable} with respect to all actions if for all intervened upon variables $X_l$ (where $l \in I^\text{total}$) the $d$-separation\footnote{Background on d-separation in Appendix \ref{appendix:d-separation}.} $\mathcal{I}^l \idp_{\mathcal{G}^*} Y | X_S$ holds in $\mathcal{G}^*$. The authors show that intervention stability implies an invariant condition distribution, i.e., for all actions $a, b \in \mathbb{A}$ with $I_a, I_b \subseteq I^{\text{total}}$ it holds that $P(Y^a|X_S) = P(Y^b|X_S)$ (\citet{pfister_stabilizing_2019}, Appendix A).
\section{Meaningful Algorithmic Recourse}
\label{sec:mar}

Algorithmic recourse searches optimal actions by assessing the prediction over a range of interventions. 
However, since predictive models are designed to be employed in a static observational distribution, they often fail to generalize to interventional environments:
For example, predictive models exploit all useful associations with the target, irrespective of the type of causal relationship between feature and target.
As a consequence, variables that are not causal for the target can be causal for the prediction \cite{Molnar2020pitfalls}. Interventions on such non-causal features may flip the prediction, but do not affect the underlying target. Thus, the model's performance is not stable under such interventions.\\
This lack of intervention stability is problematic for two reasons: Firstly, following a recourse recommendations might not lead to improvement but rather game the predictor. Secondly, a refit with access to the post-recourse distribution in which the exploited associations are weakened will not honor the original recourse recommendation.\\
The related field of strategic prediction aims to adapt the model to strategic agent behavior \cite{miller_strategic_2020}. However, the approach has a catch: As \citet{shavit_causal_2020} argue, designing the model to incentivize agent outcome (improvement) is often in conflict with achieving optimal predictive accuracy. For instance, in the example in Section \ref{sec:introduction} the model's reliance on \textit{minivan} would need to be to be reduced to incentivize improvement.\\
We propose an alternative: Instead of altering the model such that gaming is not lucrative, we allow the model to use gameable associations but constrain algorithmic recourse recommendations to those that are meaningful (Definition \ref{def:mar}). Therefore we require knowledge of the full SCM generating (observed) feature and target variables.
\begin{assumption}
We assume knowledge of the SCM that generates $X$ and $Y$. Furthermore, we assume the existence of $\mathbb{F}^{-1}(x^F)$ such that $\mathbb{F}^{-1}(x^F) = u^F$ (with $\mathbb{F}(u^F) = (x^F, y^F)$).
\end{assumption}
\begin{definition}
  Meaningful actionable recourse (MAR) is algorithmic recourse \citep{karimi_algorithmic_2021}, with the additional constraint that the underlying target $Y$ is improved coherently, i.e., $ y^{SCF,a} \geq t$ where $(y^{SCF,a}, x^{SCF,a}) = \mathbb{F}_a(\mathbb{F}^{-1}(x^F))$. \footnote{The assumption implies that an $f$ exists such that $Y = f(X)$.}
\label{def:mar}
\end{definition}
Naturally, such a restriction reduces the insight that we gain into the model. However, in our view, depending on the goal of the explainee full model insight is not required. Therefore, we distinguish two tales of algorithmic recourse in Section \ref{sec:two-tales}. Since we may not have access to the causal model required to compute MAR we propose an alternative formulation of MAR that only relies on the predictor (Section \ref{sec:robustness-mar}).

\section{The two tales of algorithmic recourse}
\label{sec:two-tales}

Machine learning explanations as suggested by \cite{karimi_algorithmic_2020,karimi_algorithmic_2021} may be used for two distinct purposes—for \textit{model audit} and for \textit{meaningful, actionable recourse}.
\\
Model auditors aim to make sure that models meet desired standards (e.g., fairness) and extrapolate well to unseen regions. Model audit explanations can allow inspectors to contest model decisions, suggest model-debugging strategies or give insight into model behavior within and outside of the data distribution \cite{Wachter2018,freiesleben2020counterfactual}. Hence, these explanations must be maximally faithful to the prediction model.
\\
Recourse recommendations on the other side need to satisfy various side constraints that are not related to the model. Even the causal dependencies between variables that are taken into account in algorithmic recourse are not reflected in the prediction model \cite{karimi_algorithmic_2021}. Recourse recommendations must also be actionable for the explainee, thus, changes in non-actionable features like age, ethnicity, or height are commonly prohibited \cite{ustun_actionable_2019,karimi_algorithmic_2021}. Moreover, recourse recommendation must be plausible, i.e., make realistic suggestions that are jointly satisfiable and prefer sparse over widespread action recommendations \cite{karimi2020model,dandl_multi-objective_2020}.
\\
In conclusion, model audit explanations are more complete and faithful to the model while recourse recommendations are more selective, faithful to the underlying process and account for the limitations of the data-subject. We believe that the selectivity and reliance of recourse recommendations on factors beside the model itself is not a limitation but essential to make explanations more relevant to the data-subject. In the same vein, we see MAR as another step towards making recourse recommendations more meaningful. 
\\
If however a data-subject is more interested in contesting or auditing the algorithmic decision, recourse recommendations are not suitable. Instead, we suggest that data-subjects should additionally receive model audit explanations upon request.

\section{Formulation based on Effective Intervention Constraint}
\label{sec:robustness-mar}

\begin{figure}
    \centering
    \begin{tikzpicture}[thick, scale=0.65, every node/.style={scale=.6, line width=0.25mm, black, fill=white}]
    \usetikzlibrary{shapes}
    
        \node[circle, draw,scale=0.9] (x1) at (-2, 1) {$X_1$};
        \node[circle, double, draw,scale=0.9] (x2) at (-1, 1) {$X_2$};
        \node[circle, draw,scale=0.9] (x3) at (-3, 0) {$X_3$};
        \node[circle, double, draw,scale=0.9] (x4) at (-2, 0) {$X_4$};
        \node[circle, double, draw,scale=0.9] (x5) at (-1, -1) {$X_5$};
        \node[circle, double, draw,scale=0.9] (x6) at (1, 1) {$X_6$};
        \node[circle, double, draw,scale=0.9] (x7) at (2, 1) {$X_7$};
        \node[circle, double, draw,scale=0.9] (x8) at (1, -1) {$X_8$};
		\node[draw,scale=1.3] (y) at (0, 0) {$Y$};

		\draw[->] (x2) -- (x1);
		\draw[->] (x2) -- (y);
		\draw[->] (x3) -- (x4);
		\draw[->] (x4) -- (y);
		\draw[->] (x5) -- (y);
		\draw[->] (y) -- (x6);
		\draw[->] (x7) -- (x6);
		\draw[->] (x8) -- (x6);
		\draw[->] (x8) -- (x5);
		

		\node[draw=orange, ellipse, scale=0.9] (i1) at (-2, 2) {$\mathcal{I}_1$};
		\node[draw=green, ellipse, scale=0.9] (i2) at (-1, 2) {$\mathcal{I}_2$};
		\node[draw=green, ellipse, scale=0.9] (i3) at (-3, -2) {$\mathcal{I}_3$};
		\node[draw=green, ellipse, scale=0.9] (i4) at (-2, -2) {$\mathcal{I}_4$};
		\node[draw=green, ellipse, scale=0.9] (i5) at (-1, -2) {$\mathcal{I}_5$};
		\node[draw=red, ellipse, scale=0.9] (i6) at (1, 2) {$\mathcal{I}_6$};
		\node[draw=orange, ellipse, scale=0.9] (i7) at (2, 2) {$\mathcal{I}_7$};
		\node[draw=green, ellipse, scale=0.9] (i8) at (1, -2) {$\mathcal{I}_8$};

		\draw[->, orange] (i1) -- (x1);
		\draw[->, green] (i2) -- (x2);
		\draw[->, green] (i3) -- (x3);
		\draw[->, green] (i4) -- (x4);
		\draw[->, green] (i5) -- (x5);
		\draw[->, red] (i6) -- (x6);
		\draw[->, orange] (i7) -- (x7);
		\draw[->, green] (i8) -- (x8);

		\draw[-, dotted] (3, -2.5) -- (3, 2.5);
		
		
		\node[circle, double, draw,scale=0.9] (x1r) at (5, 1) {$X_1$};
        \node[circle, dotted, draw,scale=0.9] (x2r) at (6, 1) {$X_2$};
        \node[circle, draw,scale=0.9] (x3r) at (4, 0) {$X_3$};
        \node[circle, double, draw,scale=0.9] (x4r) at (5, 0) {$X_4$};
        \node[circle, double, draw,scale=0.9] (x5r) at (6, -1) {$X_5$};
        \node[circle, double, draw,scale=0.9] (x6r) at (8, 1) {$X_6$};
        \node[circle, double, draw,scale=0.9] (x7r) at (9, 1) {$X_7$};
        \node[circle, dotted, draw,scale=0.9] (x8r) at (8, -1) {$X_8$};
		\node[draw,scale=1.3] (yr) at (7, 0) {$Y$};

		\draw[->] (x2r) -- (x1r);
		\draw[->] (x2r) -- (yr);
		\draw[->] (x3r) -- (x4r);
		\draw[->] (x4r) -- (yr);
		\draw[->] (x5r) -- (yr);
		\draw[->] (yr) -- (x6r);
		\draw[->] (x7r) -- (x6r);
		\draw[->] (x8r) -- (x6r);
		\draw[->] (x8r) -- (x5r);
		

		\node[draw=red, ellipse, scale=0.9] (i1r) at (5, 2) {$\mathcal{I}_1$};
		\node[draw=red, ellipse, scale=0.9] (i2r) at (6, 2) {$\mathcal{I}_2$};
		\node[draw=green, ellipse, scale=0.9] (i3r) at (4, -2) {$\mathcal{I}_3$};
		\node[draw=green, ellipse, scale=0.9] (i4r) at (5, -2) {$\mathcal{I}_4$};
		\node[draw=red, ellipse, scale=0.9] (i5r) at (6, -2) {$\mathcal{I}_5$};
		\node[draw=red, ellipse, scale=0.9] (i6r) at (8, 2) {$\mathcal{I}_6$};
		\node[draw=orange, ellipse, scale=0.9] (i7r) at (9, 2) {$\mathcal{I}_7$};
		\node[draw=red, ellipse, scale=0.9] (i8r) at (8, -2) {$\mathcal{I}_8$};

		\draw[->, red] (i1r) -- (x1r);
		\draw[->, red] (i2r) -- (x2r);
		\draw[->, green] (i3r) -- (x3r);
		\draw[->, green] (i4r) -- (x4r);
		\draw[->, red] (i5r) -- (x5r);
		\draw[->, red] (i6r) -- (x6r);
		\draw[->, orange] (i7r) -- (x7r);
		\draw[->, red] (i8r) -- (x8r);
		
    \end{tikzpicture}
    \caption{A schematic drawing illustrating under which interventions $\mathcal{I}_1, \dots, \mathcal{I}_8$ the markov blanket (double circle) is intervention stable. In this setting, we consider the intervention variables to be independent treatment variables: We would like to know how the different actions influence the conditional distribution, irrespective of how likely they are to be applied. Therefore, they are modeled as parent-less variables. Green indicates intervention stability, red indicates no intervention stability. Orange indicates intervention stability of non-causal variables. Dotted variables are not observed ($X_3$). \textit{Left:} Since all endogenous variables are observed, $MB(Y)$ is stable w.r.t. interventions on every endogenous cause of $Y$ (Proposition \ref{proposition:all-observed}). \textit{Right:} Unobserved variables open paths between interventions on causes and $Y$.}
    \label{fig:intervention-types}
\end{figure}
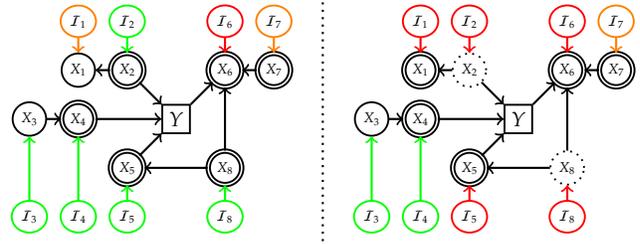

Even if we have access to the SCM modeling $X$, we may not know the structural equation generating $Y$.
Over the course of this Section we therefore introduce \textit{Effective actionable recourse} (EAR) as an alternative. Instead of computing whether the counterfacutal $y^{SCF,a}$ is flipped as desired (MAR), EAR restricts actions to exclusively intervene on causes of $Y$. Since any meaningful recourse recommendation intervenes on causes exclusively, the constraint does not exclude any $MAR$ recommendation. And, as we demonstrate in this Section, access to the full causal graph $\mathcal{G}$ can suffice to ensure robustness of the model $f$ to interventions on causes of $Y$ and therefore meaningful recourse.
\begin{definition}
  Effective actionable recourse (EAR) is algorithmic recourse \citep{karimi_algorithmic_2021} with the further constraint that only effective actions $a \in \mathcal{E}\text{ffective}$ are allowed, i.e. that $I_a \subseteq \mathcal{C}\text{auses}(Y)$
\label{def:car}
\end{definition}
We leverage research on invariant prediction \cite{pfister_stabilizing_2019} (Section \ref{sec:background-stability}) to formalize under which assumptions EAR provides meaningful recourse. For the recommendation to lead to improvement, the model has to predict accurately in the respective action distribution. Since intervention stability implies invariant conditionals (Section \ref{sec:background-stability}), for predictors that rely on intervention stable sets, EAR recommendations are as likely to lead to improvement as the predictor is able to correctly predict in the pre-recourse distribution. \\
In Figure \ref{fig:intervention-types} we illustrate under which interventions the minimal optimal set, the so-called Markov blanket, is intervention stable. Thereby, the set of observed variables $O$ plays a crucial role. For example, if the minimal set of variables in $\mathcal{G}$ that $d$-separates all remaining variables from $Y$ is observed, the markov blanket is intervention stable (Proposition \ref{proposition:all-observed}). In contrast, an intervention on unobserved direct causes (e.g., $X_2$) (as well as interventions on non-causal variables) may alter the conditional $P(Y|MB(Y))$.
\begin{proposition}
If all endogenous direct causes, direct children and spouses are observed, the markov blanket is stable with respect to interventions on all endogenous causes of $Y$.
\label{proposition:all-observed}
\end{proposition}
If the predictor is perfect and intervention stable, meaning that $f(x) = y$, EAR affects prediction and target coherently.
\begin{proposition}
Assuming a perfect predictor $f$ (i.e., $f(x) = y$ in the pre-recourse distribution) and that the predictor relies on a variable set $S$ that is stable with respect to interventions on actionable causes of $Y$, for EAR recommendations with $P(X=x^{SCF}_S)>0$ it holds that $\hat{y}^{SCF} = y^{SCF}$.
\label{proposition:car-improvement}
\end{proposition}
Even if the conditional distribution is invariant, the joint distribution of the variables is most certainly affected if individuals act on recourse recommendations. If the interventional joint distribution extends the support of the observation distribution, the model would be forced to predict outside of the training distribution. EAR recommendations that extrapolate should therefore be handled with care.\\
If the conditional is stable and the predictor is able to perfectly predict $Y$ under any action that may be recommended by recourse, ceteris paribus the recourse will be honored by an optimal refit of the model on pre- and post-recourse data.

\section{Limitations and Discussion}

In order to generate more robust and meaningful recourse recommendations affecting not only $\hat{Y}$ but also $Y$, we introduce meaningful algorithmic recourse (MAR) and a relaxation called effective algorithmic recourse (EAR).\\
Our approach is based on strong assumptions: We require that $Y$ can be perfectly predicted from features and assume existence and knowledge of an underlying SCM with invertible structural equations. EAR requires that the model is stable w.r.t. interventions on causes. Evaluating the intervention stability requires knowledge of the causal graph.\\
Furthermore, one may argue that explanations should be maximally faithful to the model. Then, the gameability should be exposed. However, we argue that recommending gaming is problematic for both model authorities \textit{and} individuals who seek robust recommendations that help them to improve. To reconcile both positions, we recommend offering additional model audit explanations upon request, but to only offer guarantees for MAR. In order to provide a variety of options, model authorities should aim to observe causes rather than non-causal variables. If the model is able to predict accurately based on causes alone, model audit and meaningful, actionable recourse converge.\\
It could be insisted that even if the conditional is invariant to MAR, the actions induce a shift in the distribution of Y. Indeed, if the threshold acts as a gatekeeper towards a limited good, the threshold itself may shift as a result. Consequently, even if the underlying target Y improves as desired, recourse may not be honored.\\
Further research is required to transfer the results into a probabilistic setting with imperfect causal knowledge. In such settings the robustness of recourse with respect to refits is particularly challenging since recourse may be applied selectively and thereby can open additional dependence paths that allow for an improved but different predictor.\\
We see our work as a first step towards more meaningful and robust recourse.


\section*{Acknowledgments}

This project is funded by the German Federal Ministry of Education and Research (BMBF) under Grant No. 01IS18036A and the Graduate School of Systemic Neurosciences (GSN) Munich.

\bibliography{main}
\bibliographystyle{icml2021}

\appendix

\section{Proofs}
\label{appendix:proofs}

\subsection{Proof of Proposition \ref{proposition:all-observed}}

\textit{If all endogenous direct causes, direct children and spouses are observed, the conditional $P(Y|MB(Y))$ is stable with respect to interventions on any set of endogenous causes of $Y$.
}\\
We prove the statement in four steps.\\
\\
\textit{Given a graph $\mathcal{G}$ and an endogenous $Y$, the set of endogeneous direct parents, direct effects and direct parents of effects are the minimal $d$-separating set $S_{\mathcal{G}}$:} Standard result, see e.g. \citet{Peters2017book}, Proposition 6.27.\\ 
\textit{The set $S_{\mathcal{G}^*}$ in the augmented graph $\mathcal{G}^*$ coincides with $S_{\mathcal{G}}$:} The minimal $d$-separating set contains direct causes, direct effects and direct parents of direct effects. $\mathcal{I}_l$ is never a direct cause of $X_l$. Also, since $\mathcal{I}_l$ has no endogenous causes, it cannot be a direct effect. Furthermore, since we restrict interventions to be performed on causes, $\mathcal{I}_l$ cannot be a direct parent of a direct effect.\\
\textit{$S_\mathcal{G}$ is intervention stable:} As follows, all intervention variables are $d$-separated from $Y$ in $\mathcal{G^*}$ by $S_\mathcal{G}$. Therefore $S_{\mathcal{G}}$ is intervention stable.\\
\textit{Then also the markov blanket is intervention stable:} 
Since $d$-separation implies independence $MB(Y) \subseteq S_\mathcal{G}$. Therefore, if $X_T \idp Y | MB(Y)$ then also $X_T \idp Y | S_{\mathcal{G}}$. If any element $s \in S_\mathcal{G}$ it holds that $s \not \in MB(Y)$, then it must hold that $X_s \idp Y | MB(Y)$. Therefore, if $X_T \idp Y | MB(Y), X_s$ then also $X_T \idp Y | MB(Y)$ and therefore any independence entailed by $S_\mathcal{G}$ also holds for $MB(Y)$. Since \cite{pfister_stabilizing_2019} only require the independence that is implied by $d$-separation in their invariant conditional proof, the same implication holds for the $MB(Y)$. As follows, $P(Y|MB(Y))$ is invariant with respect to interventions on any set of endogenous causes.\\

\subsection{Proof of Proposition \ref{proposition:car-improvement}}

\textit{Assuming a perfect predictor $f$ (i.e. $f(X) = Y$ in the pre-recourse distribution) and that the predictor relies on a variable set $S$ that is stable with respect to interventions on actionable causes of $Y$, for EAR recommendations with $P(X=x^{SCF}_S)>0$ it holds that $\hat{y}^{SCF} = y^{SCF}$.}\\
Since $S$ is intervention stable with respect to all actions $a$, for any two actions $a, b$ it holds that $P(Y^a|X_S) = P(Y^b|X_S)$\footnote{For a proof please refer to \cite{pfister_stabilizing_2019}, Appendix A.} and therefore if $y^a \neq y^b \Rightarrow x_S^a \neq x_S^b$ . Since the predictor works perfectly under the null-action $c$ (no interventions), $f(X_S) = y^c = y^a$ if $P^c(x^S) > 0$, which proves the statement.\\

\section{Background}
\subsection{$d$-separation}
\label{appendix:d-separation}

Two variable sets $X, Y$ are called $d$-separated \cite{geiger1990identifying,Spirtes2001} by the variable set $Z$ in a graph $\mathcal{G}$ ($X \idp_{\mathcal{G}} Y | Z$), if, and only if, for every path $p$ holds either (i) $p$ contains a chain $i \rightarrow m \rightarrow j$ or a fork $i \leftarrow m \rightarrow j$ where $m \in Z$ or (ii) $p$ contains a collider $i \rightarrow m \leftarrow j$ such that $m$ and all of its descendants $n$ it holds that $m, n \not \in Z$.

\end{document}